\documentclass{article}
\usepackage{ijcai23}
\usepackage{times}
\usepackage{soul}
\usepackage{url}
\usepackage[hidelinks]{hyperref}
\usepackage[utf8]{inputenc}
\usepackage[small]{caption}
\usepackage{graphicx}
\usepackage{amsmath}
\usepackage{amsthm}
\usepackage{booktabs}
\usepackage{algorithm}
\usepackage{algorithmic}
\usepackage[switch]{lineno}
\usepackage{amssymb}
\usepackage{multirow}
\usepackage{wrapfig}
\usepackage{marvosym}
\usepackage{makecell} 
\usepackage{colortbl}
\usepackage{color,xcolor}
\usepackage{subcaption}

\newcommand{\ve}[1]{\mathbf{#1}} 

\usepackage[capitalize]{cleveref}
\crefname{section}{Sec.}{Secs.}
\Crefname{section}{Section}{Sections}
\Crefname{table}{Table}{Tables}
\crefname{table}{Tab.}{Tabs.}

\title{Teaching What You Should Teach: A Data-Based Distillation Method}

\author{
Shitong Shao$^1$\and
Huanran Chen$^1$\and
Zhen Huang$^2$\and
Linrui Gong$^3$\and
Shuai Wang$^4$\And
Xinxiao Wu$^{1}$\footnote{Corresponding author}
\affiliations
$^1$Beijing Key Laboratory of Intelligent Information Technology, School of Computer Science and Technology, Beijing Institute of Technology, China\\
$^2$Univeristy of Science and Technology of China, Hefei, China\\
$^3$Hunan University, Hunan, China\\
$^4$Tsinghua University, Beijing, China \\
\emails
1090784053sst@gmail.com,
huanranchen@bit.edu.cn,
zhenhuang@mail.ustc.edu.cn,
linruigong965@gmail.com,
s-wang20@mails.tsinghua.edu.cn,
wuxinxiao@bit.edu.cn
}

\begin{document}
\maketitle


\begin{abstract}
In real teaching scenarios, an excellent teacher always teaches what he (or she) is good at but the student is not. This gives the student the best assistance in making up for his (or her) weaknesses and becoming a good one overall. Enlightened by this, we introduce the ``Teaching what you Should Teach'' strategy into a knowledge distillation framework, and propose a data-based distillation method named ``TST'' that searches for desirable augmented samples to assist in distilling more efficiently and rationally. To be specific, we design a neural network-based data augmentation module with priori bias to find out what meets the teacher's strengths but the student's weaknesses, by learning magnitudes and probabilities to generate suitable data samples. By training the data augmentation module and the generalized distillation paradigm alternately, a student model is learned with excellent generalization ability. To verify the effectiveness of our method, we conducted extensive comparative experiments on object recognition, detection, and segmentation tasks. The results on the CIFAR-100, ImageNet-1k, MS-COCO, and Cityscapes datasets demonstrate that our method achieves state-of-the-art performance on almost all teacher-student pairs. Furthermore, we conduct visualization studies to explore what magnitudes and probabilities are needed for the distillation process.
\end{abstract}
\section{Introduction}
\label{sec:intro}
Deep neural networks have been widely applied in many fields, such as computer vision~\cite{Zagoruyko2016WRN,centreidloss,chenhuanran_1}, natural language processing~\cite{transformer,da_transformerxl_2019}, reinforcement learning~\cite{vo_dqn_2013} and speech signal processing~\cite{pu_asp_2019}. However, the improving performance of the deep learning model is always achieved by the increasing size of the model, which makes it impractical to deploy the large-scale model further in real-world scenarios, especially on small devices~\cite{he_mae_2022,ja_bert_2018}. To alleviate this problem, researchers have developed a series of model compression methods, such as parameter pruning~\cite{parameterpruning}, quantization~\cite{wu_quantized_2016}, and knowledge distillation~\cite{vanillakd}.

Inspired by the teaching process in the human world, knowledge distillation is firstly proposed by~\cite{vanillakd}, which aims to improve the target model or so-called student model, by teaching this model using a better teacher model instead of only ground truth labels. There are two mainstream approaches to distilling knowledge from the teacher to the student. One is the logit-based distillation, where the student is supervised not only by the ground truth labels but also the output of the teacher~\cite{DKD,DIST}. The other is the feature-based distillation, which not only aligns the output of the student and the teacher, but also aligns the activation map in some layers~\cite{SPKD,VID}. These methods have achieved remarkable progress on learning efficient and effective models in the past decades, and the theoretical and practical systems of these two research directions are pretty well established. Thus, it is difficult for scholars to come up with new algorithms to surpass the past ones and further improve the student's performance.

In this work, we rethink the knowledge distillation algorithm from a new perspective called data-based distillation. In a real-world teaching scenario, if logit-based distillation and feature-based distillation can be treated as teaching in different styles and ways, then data-based distillation can be seen as teaching knowledge in different fields. Commonly, a good teacher will decide which subject to teach based on the student's own deficiencies, and, of course, the teacher should at least be proficient in that subject. This gives the student the knowledge it most urgently needs and drives it to improve its abilities in the corresponding weak subject. Thus, we can introduce this teaching idea to knowledge distillation and let the teacher only teach the knowledge that it should teach to boost the performance and effectiveness of the student.
\begin{figure*}[t]
\includegraphics[width=1.0\textwidth,trim={0cm 0cm 7cm 0cm},clip]{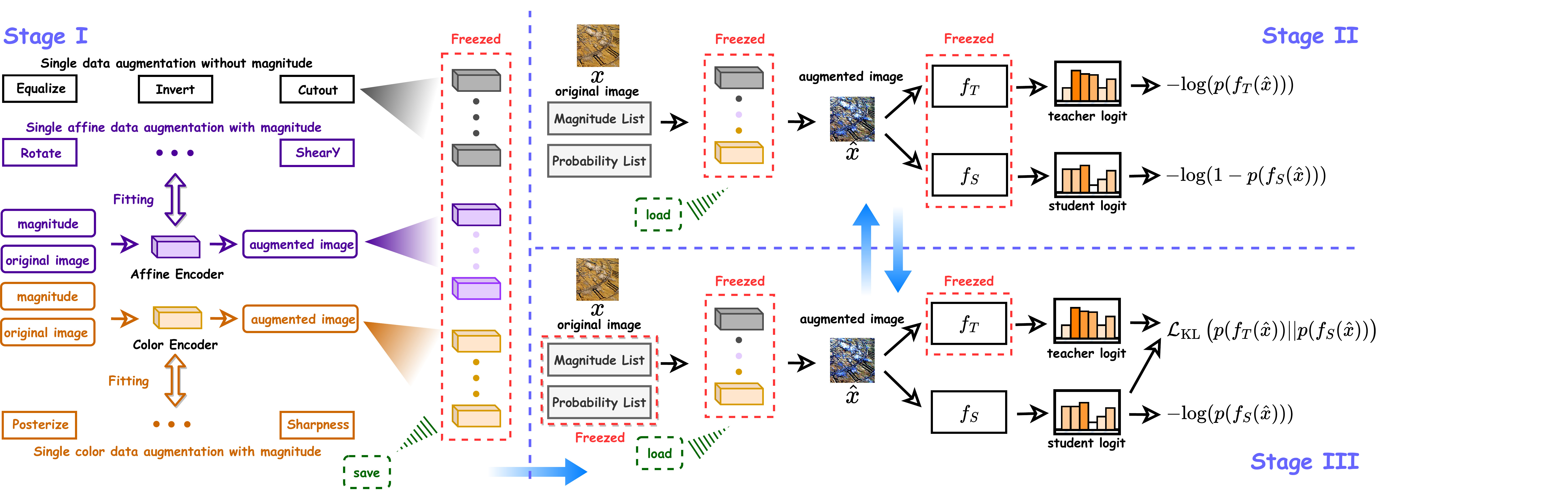}
\caption{The overall framework of TST. In Stage I, we introduce priori bias about data augmentation into the neural network-based data augmentation module. Then, we switch between updating magnitudes and probabilities in Stage II and $f_S$ in Stage III to make up for as many of the weaknesses of the student as possible.}
\label{fig:sdakd_overall_structure}
\end{figure*}

More specifically, we propose a novel data-based distillation method, named ``Teaching what you Should Teach (TST),'' which can improve the student's generalization ability by generating a series of new samples that the teacher excels at but the student does not. We design a mechanism that contains a data augmentation encoder that aims to generate augmented samples that the teacher is good at while the student is not good at; then, the teacher utilizes these augmented samples to guide the student. Note that this encoder is differentiable and stable, and the safety and diversity of generated samples can be ensured since it introduces manual data augmentations (e.g., Translate, Rotate, and Solarize) and employs the microscopic property of meta-encoders. In addition, we assume that it is easy for TST to generate samples that meet its requirements, which is difficult for the student to learn. Thus, we make the number of training iterations of the encoder much smaller than that of the student to ensure the distilled students have the most robust performance improvement.

We conduct extensive experiments on image classification (CIFAR-100~\cite{CIFAR} and ImageNet-1k~\cite{ILSVRC15}), object detection (MS-COCO~\cite{COCO}), and semantic segmentation (Cityscapes~\cite{cordts2016cityscapes}) tasks. As a result, TST achieves state-of-the-art (SOTA) performance in all quantitative comparison experiments with fair comparison, indicating that data-based distillation is feasible and needs more scholarly attention. 

Our contribution can be summarized as follows:
\begin{itemize}
\setlength{\itemsep}{0pt}
\setlength{\parsep}{0pt}
\setlength{\parskip}{0pt}
        \item[$\bullet$] We rethink knowledge distillation from a data perspective, which not only inspires us to draw on real-world teaching scenarios but also helps us come up with a new idea, improving knowledge distillation by teaching methods as well as using more desirable data.
        \item[$\bullet$] We propose a novel data-based distillation method named TST to help search for augmented samples that are suitable for distillation. In TST, we develop a differentiable and stable data augmentation encoder to generate more data like that and add priori bias into this encoder to ensure its stability and interpretability.
        \item[$\bullet$] We have achieved SOTA performance on a range of vision tasks and done plenty of ablation studies to verify the effectiveness of TST. Besides, we visualize the characteristic information of the data augmentation encoder learned by TST and discover some patterns in it.
\end{itemize}

\section{Related Work}
\label{sec:related_work}
\paragraph{Categorize Knowledge Distillation.} We classify knowledge distillation algorithms into three categories: logit-based distillation, feature-based distillation, and data-based distillation. Past researchers have explored knowledge distillation only at the logit or feature levels. Specifically, vanilla KD~\cite{vanillakd}, DKD~\cite{DKD}, and DIST~\cite{DIST} measure the difference in probability distribution between the student's logit and the teacher's logit by designing different forms of functions. And FitNet~\cite{FitNet}, ATKD~\cite{ATKD}, and CRD~\cite{CRD} apply encoders or novel distance metric functions, or both to perform knowledge transfer at the feature level. However, these works focus only on ``how to teach'' and ignore ``teach what'', which can be analogized to the data level. We categorize knowledge distillation algorithms on the data level as data-based distillation and find it has not been investigated in the generic distillation framework~\cite{data_free_yin2020dreaming,data_free_fang2022up}. To fill this gap, we turn to focus on designing a data-based distillation method and use a data augmentation encoder to obtain new samples from the original samples that can be recognized by the teacher but not by the student.

\paragraph{Learnable Data Augmentation.} Data augmentation utilizes affine and color transformations to expand the training dataset. Traditional manual data augmentation, including RandAugment~\cite{RandAugment} and AutoAugment~\cite{AutoAugment}, is non-differentiable and unlearnable. In recent years, differentiable data augmentation~\cite{DADA_ECCV_2020} has ensured the learnability of data augmentation through a series of numerical approximations, but its overly complex framework design has hindered its further application. TeachAugment~\cite{suzuki2022teachaugment} effectively solves the problem of unlearnability of data augmentation through a neural network-based data augmentation module, but the instability of this module leads to the need to impose a series of additional constraints to assist training. Thus, in TST, we add priori bias to the neural network-based data augmentation module and then freeze it in standard training. This ensures that the new data encoder is interpretable and stable.

\section{Teach What You Should Teach}
In this section, we first introduce the total framework of TST. Then, we present the simple knowledge distillation in Stage III, the search for desirable data in Stage II, and how to introduce priori bias into the data augmentation module in Stage I. Finally, the concrete implementation of the data augmentation module is described.
\subsection{Total Framework}
As illustrated in Fig.~\ref{fig:sdakd_overall_structure}, TST can be divided into three stages, namely Stage I, Stage II, and Stage III. To be specific, Stage I introduces the reasonable priori bias into the neural network-based data augmentation module before formal training, which helps ensure the stability of training in Stage II and Stage III, and improves the student's ultimate generalization ability. Then, Stages II and III obtain an excellent student by alternately training them. In Stage II, we initialize the learnable magnitudes and probabilities, and search augmented samples that the teacher is good at but the student is not, by minimizing the cross-entropy of the teacher and maximizing the cross-entropy of the student. We expect that the student will serve its deficiencies in Stage III and obtain good generalization ability after training. In particular, Stage III can be viewed as a simple knowledge distillation framework that only applies a traditional cross-entropy loss and a Kullback-Leibler divergence to optimize the student. Therefore, the core of TST is to find more desirable training data for Stage III through Stage I and Stage II. The more detailed algorithmic procedure of TST can be found in Appendix~\ref{apd:TST_procedure}.

\subsection{Simple Knowledge Distillation}
Stage III is a simple knowledge distillation framework, which can be interpreted as follows: given a teacher $f_T$ with parameter $\theta_T$ and a student $f_S$ with parameter $\theta_S$, when a training set $\mathcal{X}$ including the original samples and augmented samples are utilized for training, we can sample the mini-batch $\{x^k\}^B_{k=1}$ ($B$ denotes the batch size) from it and get the student output $\{f_S(x^k)\}^B_{k=1}$ and the teacher output $\{f_T(x^k)\}^B_{k=1}$ by forward propagation. Let us define the normalized exponential function $p$, i.e., $\textrm{softmax}$, to calculate the probability distribution of the model output.
The goal of knowledge distillation is to minimize the cross-entropy loss $\mathcal{L}_\textrm{CE}$ between $\{p(f_S(x^k))\}^B_{k=1}$ and the ground truth label $\{y^k\}^B_{k=1}$, and the Kullback-Leibler divergence $\mathcal{L}_\textrm{KL}$ between $\{p(f_T(x^k)/\tau)\}^B_{k=1}$ and $\{p(f_S(x^k)/\tau)\}^B_{k=1}$, where $\tau$ refers to the temperature weight. Thus, the total loss function $\mathcal{L}_\textrm{total}$ is formulated as
\begin{equation}
\small
\begin{aligned}
\mathcal{L}_\textrm{total}\!=\!\frac{1}{B}\sum_{k=1}^B&w_\textrm{ce}\mathcal{L}_\textrm{CE}\left(p(f_S(x^k)),y^k\right)\\
&+\!w_\textrm{kl}\tau^2\mathcal{L}_\textrm{KL}\left(p(f_T(x^k)/\tau)||p(f_S(x^k)/\tau)\right), \\
\end{aligned}
\label{eq:totalloss}
\end{equation}
where $w_\textrm{ce}$ and $w_\textrm{kl}$ are balanced weights. Note that Eq.~\ref{eq:totalloss} can also be appended with other distillation loss functions, e.g. Mean Squared Error function ($\mathcal{L}_{MSE}$), to improve the performance of distillation, which is applied in TST for the object detection task.

\subsection{Search Desirable Data}
The goal of TST is to assist the student in overcoming its deficiencies in Stage III by discovering more desirable data that the teacher excels at but the student does not in Stage II. We denote the neural network-based data augmentation module $f_{DE}$ contains a data augmentation meta-encoder set $\{f^i_E\}_{i=1}^N$ with a parameter set $\{\theta^i_E,\theta^i_p\}_{i=1}^N\cup\{\theta^i_m\}_{i=1}^{N-N_\textrm{nl}}$, where $N$ and $N_\textrm{nl}$ refer to the total number of all sub-policies and the total number of sub-policies without magnitudes, respectively. Note that each $f_E$ has the frozen parameter $\theta_E$ containing priori bias, and its other parameters, i.e., the magnitude\footnote{Magnitudes indicate the strength of the sub-policies transition. Some sub-policies do not have this variable, such as Cutout, Equalize, and Invert. These details will be introduced later.} $\theta_m$ and the probability $\theta_p$, are learnable in Stage II. Therefore, as shown in the upper right corner of Fig.~\ref{fig:sdakd_overall_structure}, we denote the loss function of TST in Stage II as
\begin{equation}
\small
\begin{aligned}
\mathcal{L}_\textrm{TST}=\frac{1}{B}\sum_{k=1}^B&\alpha\mathcal{L}_\textrm{CE}\left(p(f_T(\hat{x}^k)),y^k\right)\\
&+\!\beta\mathcal{L}_\textrm{CE}\left(1\!-\!p(f_S(\hat{x}^k)),y^k\right), \\
\end{aligned}
\label{eq:TSTloss}
\end{equation}
where $\hat{x}$ denotes the augmented sample obtained from the original sample after $f_{DE}$, and $\alpha$ and $\beta$ denote the balanced weights. Specific details about $f_{DE}$ will be presented in Sec.~\ref{sec:DE}. The optimization objective of Stage II $\mathop{\textrm{min}}\limits_{\{\theta^i_m,\theta^i_p\}_{i=1}^N}\!\mathcal{L}_\textrm{TST}$ is to let the student misclassify but let the teacher correctly classify so that the teacher can transfer the most helpful knowledge to students. To better find desirable data that the student cannot classify correctly, we use $\mathcal{L}_\textrm{CE}\left(1\!-\!p(f_S(\hat{x}^k)),y^k\right)$ instead of $-\mathcal{L}_\textrm{CE}\left(p(f_S(\hat{x}^k)),y^k\right)$ to avoid non-convex optimization, and the proof can be found in Appendix~\ref{apd:sd}. Moreover, the parameters $\{\theta^i_E\}_{i=1}^N$ are frozen because they contain  priori bias in favor of distillation, which is imported in Stage I. As demonstrated later in Sec.~\ref{sec:ablation}, distillation can easily fail if $\{\theta^i_E\}_{i=1}^N$ is not frozen due to the instability of the data augmentation module. 

\begin{table*}[thp]
\center
\begin{small}
\resizebox{1.\textwidth}{!}{%
\begin{tabular}{ll|cccccc|ccc}
\multicolumn{2}{c|}{Architecture} & \multicolumn{6}{c|}{Same} & \multicolumn{3}{c}{Different} \\\hline
\multirow{4}{*}{\begin{tabular}[c]{@{}c@{}}Distillation \\ Type\end{tabular}} & \multirow{2}{*}{Teacher}  & ResNet110 & ResNet110 & WRN-40-2 & WRN-40-2 & ResNet32$\times$4 & VGG13 & WRN40-2 & ResNet32$\times$4  & VGG13\\
&      & 74.31     & 74.31     & 75.61      & 75.61      &  79.42    & 74.64 & 75.61 & 79.42 & 75.61  \\
& \multirow{2}{*}{Student}  & ResNet20~\shortcite{ResNet} & ResNet32 & WRN-40-1~\shortcite{Zagoruyko2016WRN}  & WRN-16-2 & ResNet8$\times$4 & VGG8 & ShuffleNet-V1 & ShuffleNet-V1 & MobileNet-V2 \\
& \space    & 69.06      & 71.14       & 71.98       & 73.26      & 72.50     & 70.36  &  70.50 & 70.50 & 64.60 \\ \hline
\multirow{9}{*}{Feature-based}
& FitNet~\shortcite{FitNet} &  68.99 & 71.06   & 72.24      & 73.58      & 73.50    & 71.02 & 73.73 & 73.59 &  64.14 \\
& ATKD~\shortcite{ATKD} & 70.22 & 70.55 & 72.77 & 74.08 & 73.44 & 71.43 & 73.32 & 72.73 & 59.40 \\
& SPKD~\shortcite{SPKD} & 70.04 & 72.69 & 72.43 & 73.83 & 72.94 & 72.68 & 74.52 & 73.48 & 66.30\\
& CCKD~\shortcite{CCKD} & 69.48 & 71.48 & 72.21 & 73.56 & 72.97 & 70.71 & 71.38 & 71.14 & 64.86 \\
& RKD~\shortcite{RKD} & 69.25  & 71.82 &  72.22 & 73.35 & 71.90 & 71.48 & 72.21 & 72.28 & 64.52 \\
& VID~\shortcite{VID} & 70.16 & 70.38 & 73.30 & 74.11 & 73.09 & 71.23 & 73.61 & 73.38 & 65.56 \\
& CRD~\shortcite{CRD}   & 71.46      & 73.48       & 74.14       & 75.48      & 75.51      & 73.94  & 76.05 & 75.11 & 69.73 \\
& OFD~\shortcite{COFD}   &  -   & 73.23    & 74.33   & 75.24 & 74.95 & 73.95  & 75.85 & 75.98 & 69.48\\
& ReviewKD~\shortcite{ReviewKD} &  - & 71.89 & 75.09 & 76.12 & 75.63 & 74.84 & 77.14 & \textbf{77.45} & 70.37 \\ \hline 
\multirow{3}{*}{Logit-based}       
& KD~\shortcite{vanillakd} & 70.67 & 73.08 & 73.54 & 74.92 & 73.33 & 72.98 & 74.83 & 74.07 & 67.37 \\
& DKD~\shortcite{DKD} & - & 74.11 & 74.81 & 76.24 & 76.32 & 74.68 & 76.70 & 76.45 & 69.71 \\
& DIST~\shortcite{DIST} & 69.94 & 73.55 & 74.42 & 75.29 & 75.79 & 73.74 & 75.23 & 75.23 & 68.48  \\\hline
\multirow{1}{*}{Data-based}
& Ours & \textbf{72.44} & \textbf{75.04} & \textbf{75.32} & \textbf{76.75} & \textbf{76.72} & \textbf{75.03} & \textbf{77.38} & 76.71 & \textbf{70.82} \\
\end{tabular}}
\end{small}
\caption{Results on the CIFAR-100 test set. ``Same'' and ``Different'' in the first row refer to whether the model architecture is the same for teachers and students.}
\label{tab:cifarcompareresult}
\end{table*}
\begin{table*}[htp]
\renewcommand\arraystretch{0.95}
\setlength\tabcolsep{0.6mm}
	\center
 \footnotesize
\resizebox{1.\textwidth}{!}{%
\begin{tabular}{cc|c|cc|ccccc|ccc|c}
\multicolumn{2}{c|}{Architecture} & & \multicolumn{2}{c|}{Accuracy}  & \multicolumn{5}{c|}{Feature-based} & \multicolumn{3}{c|}{Logit-based} & \multicolumn{1}{c}{Data-based} \\\hline
Teacher & Student &  & Teacher & Student & OFD~\shortcite{COFD} & RKD~\shortcite{RKD} & CRD~\shortcite{CRD} & SRRL~\shortcite{SRRL} & ReviewKD~\shortcite{ReviewKD} & KD~\shortcite{vanillakd} & DKD~\shortcite{DKD} & DIST~\shortcite{DIST} & Ours \\\hline
\multirow{2}{*}{ResNet-34} & \multirow{2}{*}{ResNet-18} & Top-1 & 73.31 & 69.76 & 71.08 & 70.34 & 71.17 & 71.73 & 71.61  & 70.66 & 71.70 & 72.07 &  \textbf{72.22} \\
		~ & & Top-5 & 91.42 & 89.08 & 90.07 & 90.37 & 90.13 & 90.60 & 90.51 & 89.88 & 90.41 & 90.42 & \textbf{90.68} \\
		\hline
	    \multirow{2}{*}{ResNet-50} & \multirow{2}{*}{MobileNet-V1} & Top-1 & 76.16 & 70.13 & 71.25 & - & 71.37 & 72.49 & 72.56 & 70.68 & 72.05 & \textbf{73.24} & 72.31\\
		~&  & Top-5 & 92.86 & 89.49 & 90.34 & - & 90.41 & 90.92 & 91.00 & 90.30 & 91.05 & \textbf{91.12} & 90.70\\\hline
\multirow{1}{*}{Swin-Large} & \multirow{1}{*}{Swin-Tiny} & Top-1 & 86.30 & 81.30 & - & 81.20 & - & 81.50 & - & 81.50 & - & \textbf{82.26} & 82.21 \\ \end{tabular}}
\caption{Results on the ImageNet validation set. We use ResNet-34 and ResNet-50 released by Torchvision~\protect\cite{marcel2010torchvision} and Swin-Large released by~\protect\cite{SWIN-T} as our teacher's pre-training weight.}
\label{tab:imagenetcompareresult}
\end{table*}
\begin{table*}[t]
\resizebox{1.\textwidth}{!}{
\begin{tabular}{l|cccccc|cccccccl|cccccc}
 \multirow{1}{*}{T$\rightarrow$S} & \multicolumn{6}{c|}{CM RCNN-X101$\rightarrow$Faster RCNN-R50} & \multicolumn{6}{c}{RetinaNet-X101$\rightarrow$RetinaNet-R50} & &  \multirow{1}{*}{T$\rightarrow$S}
 &\multicolumn{6}{c}{FCOS-R101$\rightarrow$FCOS-R50} \\
 Type & \multicolumn{6}{c|}{\textit{Two-stage detectors}} & \multicolumn{6}{c}{\textit{One-stage detectors}} & & Type & \multicolumn{6}{c}{\textit{Anchor-free detectors}}\\
Method & AP & AP$_{50}$ & AP$_{75}$ & AP$_{S}$ & AP$_{M}$ & AP$_{L}$ & AP & AP$_{50}$ & AP$_{75}$ & AP$_{S}$ & AP$_{M}$ & AP$_{L}$ & & Method & AP & AP$_{50}$ & AP$_{75}$ & AP$_{S}$ & AP$_{M}$ & AP$_{L}$ \\\cline{0-12}\cline{15-21}
Teacher & 45.6 & 64.1 & 49.7 & 26.2 & 49.6 & 60.0 & 41.0 & 60.9 & 44.0 & 23.9 & 45.2 & 54.0 & &Teacher & 40.8 & 60.0 & 44.0 & 24.2 & 44.3 & 52.4 \\
Student & 38.4 & 59.0 & 42.0 & 21.5 & 42.1 & 50.3 & 37.4 & 56.7 & 39.6 & 20.0 & 40.7 & 49.7& & Student & 38.5 & 57.7 & 41.0 & 21.9 & 42.8 & 48.6 \\\cline{0-12}\cline{15-21}
KD~\shortcite{vanillakd} & 39.7 & 61.2 & 43.0 & 23.2 & 43.3 & 51.7 & 37.2 & 56.5 & 39.3 & 20.4 & 40.4 & 49.5 & & KD~\shortcite{vanillakd} & 39.9 & 58.4 & 42.8&23.6& 44.0 & 51.1\\
COFD~\shortcite{COFD} & 38.9 & 60.1 & 42.6 & 21.8 & 42.7 & 50.7 & 37.8 & 58.3 & 41.1 & 21.6 & 41.2 & 48.3 & & FitNet~\shortcite{FitNet} & 39.9 & 58.6 & 43.1 & 23.1 & 43.4 & 52.2 \\
FKD~\shortcite{FKD} & 41.5 & 62.2 & 45.1 & 23.5 & 45.0 & 55.3 & 39.6 & 58.8 & 42.1 & 22.7 & 43.3 & 52.5 & & GID~\shortcite{GID} &42.0 & \textbf{60.4} & \textbf{45.5} & 25.6 & 45.8 & 54.2 \\
DIST~\shortcite{DIST} & 40.4 & 61.7 & 43.8 & 23.9 & 44.6 & 52.6 & 39.8 & 59.5 & 42.5 & 22.0 & 43.7 & 53.0 & & FRS~\shortcite{FRS} &40.9 & 60.3 & 43.6 & 25.7 & 45.2 & 51.2  \\
DIST+mimic~\shortcite{DIST} & 41.8 & 62.4 & 45.6 & 23.4 & 46.1 & 55.0 & 40.1 & 59.4 & 43.0 & 23.2 & 44.0 & 53.6 & & FGD~\shortcite{FGD} & \textbf{42.1} & - & - & \textbf{27.0}& \textbf{46.0}& \textbf{54.6} \\
Ours (KD)  & 40.5 & 62.4 & 44.1 & 24.0 & 44.6 & 52.1 & 39.9 & 59.6& 42.8 & 23.3 & 43.8 & 53.3 && Ours (KD) & 40.1 & 58.3 & 43.2 & 23.9 & 44.1 & 51.6\\
Ours (KD+mimic)&\textbf{42.2} &\textbf{63.4}&\textbf{46.1}&\textbf{24.1}&\textbf{46.5}&\textbf{55.6}&\textbf{40.5}&\textbf{60.0}&\textbf{43.4}&\textbf{23.9}&\textbf{44.5}&\textbf{54.4}&&Ours (KD+mimic) & 41.0 & 60.0 & 44.3 & 24.9 & 45.0 & 51.8 \\
\end{tabular}}
\caption{Results on the COCO validation set (T$\rightarrow$S refers to the distillation from T to S). Here, the content in brackets to the right of ``Ours'' refers to the methods applied in the distillation process. In addition, CM RCNN-X101 stands for Cascade Mask RCNN-X101.}
\label{tab:detectionresult}
\end{table*}
\begin{table}[th]
	\renewcommand\arraystretch{1.1}
	\setlength\tabcolsep{3mm}
	\centering
\scriptsize
\setlength{\tabcolsep}{2.8mm}{
	\begin{tabular}{l|c|l|c}
        Method & mIoU (\%) & Method & mIoU (\%) \\        \hline
        T: PSPNet-R101 & 78.55 &  S: PSPNet-R18 & 70.09\\
        SKDD~\shortcite{liu2020structured} & 74.08 & SKDS~\shortcite{liu2019structured} & 72.70\\
        IFVD~\shortcite{wang2020intra} & 74.54 & CWD~\shortcite{CWD}$^*$ & 75.54 \\
        DIST~\shortcite{DIST} & 75.74 & Ours$^*$ & \textbf{76.55} \\ 
	\end{tabular}}
 	\caption{Results on the Cityscapes validation set. $*$: The experiments are performed based on mmrazor~\protect\cite{2021mmrazor}.}
	\label{tab:segmentation}
\end{table}

During standard distillation training, Stage II and Stage III alternate. Specifically, TST first employs Stage II to find new samples suitable for distillation, and then utilizes Stage III to improve the student's generalization ability. Inspired by~\cite{Attack_2017_towards}, we claim that it is easy to search augmented samples in Stage II that match what the teacher is good at but the student is not, while it is difficult for the student to absorb knowledge contained in the augmented samples in Stage III. Therefore, the number of iterations $n_\textrm{encoder}$ of Stage II will be much smaller than the number of iterations $n_\textrm{student}$ of Stage III. Also, Sec.~\ref{sec:ablation} illustrates that the student trained by TST will have better performance when $n_\textrm{encoder}\ll n_\textrm{student}$.

\subsection{Introduce Priori Bias into $f_{DE}$}
Noise-based generative models require expensive computational costs and suffer from training instability (more detailed can be found in Appendix~\ref{sec:gan_and_mda}). Thus, Stage I is applied to introduce the augmented priori bias into $f_{DE}$ and let $\{\theta^i_p\}_{i=1}^N\cup\{\theta^i_m\}_{i=1}^{N-N_\textrm{nl}}$ be differentiable. Thanks to priori bias, TST is able to distill a strong student in a limited number of iterations. For simplicity, we consider each meta-encoder $f_E$ in the set $\{f^i_E\}_{i=1}^N$ as a black box in this paragraph. In Stage I, since Equalize, Invert and Cutout do not have the property of magnitude, they do not need to be fitted and can be called directly in Stage II and Stage III. Of course, their probabilities are learnable and will still be optimized in Stage II. Then, we categorize a series of single data augmentations as follows: (a) magnitude-unlearnable transformations, including Equalize, Invert, and Cutout; (b) learnable affine transformations, including Rotate ShearX, ShearY, TranslateX, and TranslateY; (c) learnable color transformations, including Posterize, Solarize, Brightness, Color, Contrast, and Sharpness~\cite{AutoAugment,RandAugment}. For the learnable affine transformations and color transformations, we apply Spatial Transformer Network~\cite{STN} and Color Network for fitting, respectively. Since both types (a) and (b) are fitted in the same way, we define a set of the manual data-augmentation mappings $\{f_A^i\}_{i=1}^{N-N_\textrm{nl}}$ that includes all single data augmentations with magnitude. For each matched pair $f_A$ and $f_E$ in $\{f_A^i\}_{i=1}^{N-N_\textrm{nl}}$ and $\{f_E^i\}_{i=1}^{N-N_\textrm{nl}}$, we will optimize $\mathcal{L}_{MSE}$ by $n_\textrm{fitting}$ iterations. The loss function in Stage I can be formulated as
\begin{equation}
\small
\begin{aligned}
\mathcal{L}_\textrm{encoder}&=\frac{1}{B}\sum_{k=1}^B\|f_A(x^k,m_r)-f_E(x^k,m_r)\|^2_2, \\
\end{aligned}
\label{eq:encoderloss}
\end{equation}
where $m_r\sim\mathcal{U}(0,1)$ refers to the magnitude. $\mathcal{L}_\textrm{encoder}$ will be optimized to a very small error bound after $n_\textrm{encoder}$ iterations. This guarantees that all meta-encoders are well-trained.

\subsection{Network-Based Data Augmentation}
\label{sec:DE}
We will describe the whole process of how $x$ goes through $f_{DE}$ and finally becomes $\hat{x}$ in this sub-section. We first describe how $f_{DE}$ calls $\{f^i_E\}_{i=1}^N$ to accomplish the combination of various sub-generated samples, and then we present the details of Spatial Transformer Network~\cite{STN} and Color Network. 

After normalizing $\{\theta_m^i\}_{i=1}^{N-N_\textrm{nl}}$ and $\{\theta_p^i\}_{i=1}^{N}$ to $\left[0,1\right]$ by applying the sigmoid activation function, we employ the Relaxed
Bernoulli Distribution to sample new magnitudes and probabilities for solving the non-differentiable problem~\cite{Fastautoaugment,DADA_ECCV_2020}. The process can be formulated as
\begin{equation}
\small
\begin{aligned}
&\mathbf{\textrm{RBD}}(p)\ \ \ \ \ =\mathbf{\textrm{sigmoid}}((\textrm{log}(p)+L)/\tau_l), L\!\sim\!\mathop{Logistic}(0,1), \\
&\{\theta_m^i\}_{i=1}^{N-N_\textrm{nl}}\!=\!{\left\{\mathbf{\textrm{RBD}}\left(\mathbf{\textrm{sigmoid}}\left(\theta_m^i\right)\right)\right\}}_{i=1}^{N-N_\textrm{nl}}, \\
&\{\theta_p^i\}_{i=1}^{N}\quad\ = {\left\{\mathbf{\textrm{RBD}}\left(\mathbf{\textrm{sigmoid}}\left(\theta_p^i\right)\right)\right\}}_{i=1}^{N}, \\
\end{aligned}
\label{eq:relax_sigmoid}
\end{equation}
where $\mathop{Logistic}$ and $\tau_l$ stand for the logistic distribution and the temperature, respectively, and $\tau_l$ is set as 0.05. Then, we randomly select $N_A$ non-repeating numbers $\{Z^i\}_{i=1}^{N_{A}}$ from $\{i\}_{i=1}^{N}$. $N_A$ can be interpreted as the number of randomly sampled primitives, i.e., a larger $N_A$ means that the model is more difficult to identify the augmented samples. $N_A$ is set as 4 by default in our experiments. We obtain $\{x^i_E\}_{i=1}^{N_{A}}$ from $x$ through all meta-encoders and get the final $\hat{x}$ by simple sum denoted as
\begin{equation}
\begin{aligned}
&x^i_E\!=\begin{cases}\theta_p^{Z^i}\odot f_E(x,\theta_m^{Z^i})-\theta_p^{Z^i}\odot x, & Z^i\leq N-N_\textrm{nl}\\
\theta_p^{Z^i}\odot f_E(x)-\theta_p^{Z^i}\odot x, & Z^i>N-N_\textrm{nl}\\
\end{cases}\\
&\hat{x}\!=\!x+\sum_{i=1}^{N_A}x^i_E, \\
\end{aligned}
\label{eq:total_nbdam_forward}
\end{equation}
where $\odot$ denotes the element-wise product. For object detection tasks, the summation in Eq.~\ref{eq:total_nbdam_forward} can easily lead to the undesired result that a single target object turns into multiple target objects. Therefore, we use a separate algorithm for this special task. which is explained in Appendix~\ref{apd:sd}.

Spatial Transformer Network $f_\textrm{STN}$ and Color Network $f_\textrm{CN}$ are two different neural network-based meta-encoders. Each has parameters $\theta_E$ to store priori bias learned from Stage I. However, $f_\textrm{STN}$ and $f_\textrm{CN}$ employ different ways to complete the calculation. As shown in the next Eq., when $x$ and magnitude $m$ are fed into $f_\textrm{STN}$, $\theta_E$ (can be considered as a vector) is first concatenated with $m$ to form a new vector $\hat{\theta}_E$, and then passes through a fully connected layer $\mathcal{FC}:=\mathbb{R}^{\mathbf{\textrm{len}}(\theta_E)+1}\rightarrow\mathbb{R}^6$ to obtain the matrix $A\in\mathbb{R}^{2\times 3}$, denoted as
$$f_\textrm{STN}(x,m) = \mathbf{\mathrm{Affine}}(A,x),\ \mathbf{\mathrm{where}}\ A= \mathbf{\textrm{reshape}}(\mathcal{FC}([\theta_E,m])). $$
For $f_\textrm{CN}$, its forward propagation is composed of convolution and color transformation. So, $\theta_E$ not only has a vector $\theta_{E,V}$ but also a convolution weight $\theta_{E,C}$ to record  priori bias. When $x$ and $m$ are fed into $f_\textrm{CN}$, ${\theta}_{E,V}$ and $m$ concatenate a new vector $\hat{\theta}_{E,V}$, and we let it through a fully connected layer $\mathcal{FC}:=\mathbb{R}^{\mathbf{\textrm{len}}(\theta_{E,V})+1}\rightarrow\mathbb{R}^2$ to obtain the scale and shift parameters, i.e., $\theta_\textrm{scale}$ and $\theta_\textrm{shift}$. After that, the output of $f_\textrm{CN}$ is denoted as
$$f_\textrm{CN}(x,m) =\mathcal{C}(x\odot\left(0.5\!+\!\mathbf{\textrm{sigmoid}}(\theta_\textrm{scale})\right)+\mathbf{\textrm{sigmoid}}(\theta_\textrm{shift})\!-\!0.5), $$
where $\mathcal{C}$ refers to the convolutional layer. In particular, in our experiments, we perform $\mathbf{\textrm{RBD}}$ on $A$, $\theta_\textrm{shift}$ and $\theta_\textrm{scale}$ additionally to guarantee the diversity of data augmentation to prevent $\{\theta_m^i\}_{i=1}^{N-N_\textrm{nl}}$ and $\{\theta_p^i\}_{i=1}^{N}$ from converging to a locally optimal solution.
\section{Experiment}
\label{sec:experiment}
We conduct comparison experiments on three major tasks: image classification, object detection, and semantic segmentation. The image classification datasets include CIFAR-100~\cite{CIFAR} and ImageNet-1k~\cite{ILSVRC15}; the target detection dataset includes MS-COCO~\cite{COCO}; the semantic segmentation dataset includes Cityscapes~\cite{cordts2016cityscapes}. More details about these datasets can be found in Appendix~\ref{apd:dataset}. Besides, all the experiment results on CIFAR-100 are the average over five trials, while the related experiment results on other datasets are the average over three trials. We apply batch size 128 and initial learning rate 0.1 on CIFAR-100. And we follow the settings in~\cite{DIST} for the ResNet34-ResNet18 pair and the ResNet50-MobileNet pair on ImageNet-1k. The settings of other classification, detection and segmentation tasks can be found in Appendix~\ref{apd:hsettings}.

\subsection{Comparison with SOTA Methods}
\paragraph{Classification on CIFAR-100.} We compare many state-of-the-art feature-based and logit-based distillation algorithms on nine student-teacher pairs. For these teacher-student pairs, the teacher and student of six pairs have the same structure, and the teacher and student of the other three pairs have different architecture. The experimental results are presented in Table~\ref{tab:cifarcompareresult}. Obviously, we can find that TST, as the only data-based distillation approach, outperforms all other algorithms on eight student-teacher pairs except for ResNet32$\times$4-ShuffleNet-V1. Especially on three teacher-student pairs, including ResNet110-ResNet20, ResNet110-ResNet32 and VGG13-VGG8, our TST surpasses the latest state-of-the-art methods by almost more than one percent. Besides, in order to more fully demonstrate the excellent performance of TST, we conduct simulations of few-shot scenarios on CIFAR-100. Here, we follow the training settings in~\cite{CRD} and randomly discard 25\%, 50\%, and 75\% samples for training. As the experimental results shown in Appendix~\ref{apd:few_shot_experiment}, TST also performs well in this few-shot scenario.

\paragraph{Classification on ImageNet-1k.} To further demonstrate whether TST can work robustly on ImageNet-1k, we conduct experiments with two different architecture pairs, including Conv-Conv and ViT-ViT pairs. For Conv-Conv pair, we consider two teacher-student pairs: ResNet34-ResNet18 and ResNet50-MobileNet-V1 pairs, and apply the same hyperparameter settings to show the effectiveness of TST. The results are illustrated in Table~\ref{tab:imagenetcompareresult}. We can find that TST beats all state-of-the-art methods in ResNet34-ResNet18 pair but is slightly inferior to SRRL, ReviewKD and DIST in ResNet50-MobileNet-V1 pair. The possible reason is that the teacher is stronger, which causes the teacher overconfident in discriminating on the original and generating augmented samples. In our discussion, we consider this phenomenon in isolation and give two solutions for mitigating poor student performance under TST training with the stronger teacher. In particular, this is not a defect of TST, but rather due to the fact that $\mathcal{L}_{KL}$ is not adapted to the distillation process with the stronger teacher. In fact, there are many approaches, including DIST~\cite{DIST} and DKD~\cite{DKD}, that have been proposed to alleviate the gap between the teacher and student. For the ViT-ViT pair, we regard Swin Transformer~\cite{SWIN-T}, which is widely known and applied by researchers, as the model architecture in distillation. We treat Swin Transformer Tiny (Swin-Tiny) as the student and Swin-Transformer Large (Swin-Large) as the teacher. The experimental results are presented in Table~\ref{tab:imagenetcompareresult}, where TST surpasses all methods except DIST, showing that TST is applicable to ViT-based architectures.

\paragraph{Detection on MS-COCO.} Comparison experiments are run on three kinds of different detectors, i.e., \textit{two-Stage detectors, one-stage detectors, anchor-free detectors}. In particular, TST introduces additional losses on Stage II that drove the student box regression to be inaccurate and the teacher box regression to be accurate. As shown in Table~\ref{tab:detectionresult}, TST breaks the vanilla KD bottleneck by locating augmented samples that are conducive to distillation. Due to the additional loss on box regression in Stage II and the fact that the detection task depends more on the network's ability to produce good features, we believe that aligning the student and teacher feature maps will improve the performance of TST. Thus, we follow~\cite{DIST} by adding auxiliary loss mimic, i.e., translating the student feature map to the teacher feature map by a convolution layer and supervising them utilizing $\mathcal{L}_{MSE}$, to the detection distillation task. Ultimately, we can conclude from Table~\ref{tab:detectionresult} that TST based on the vanilla KD and mimic achieves the best performance on Cascade RCNN-X101-Cascade RCNN-R50 and RetinaNet-X101-RetinaNet-R50 pairs.

\paragraph{Segmentation on Cityscapes.} We conduct comparative experiments of semantic segmentation on PSPNet-R101-PSPNet-R18 pair~\cite{pspnet}. As shown in Table~\ref{tab:segmentation}, TST outperforms all the state-of-the-art methods and demonstrates that data-based distillation is effective in the field of semantic segmentation.

\subsection{Ablation Study}
\label{sec:ablation}
\begin{table}[t]
	\renewcommand\arraystretch{1.1}
	\setlength\tabcolsep{0.6mm}
	\centering
\scriptsize
\setlength{\tabcolsep}{2.8mm}{
\begin{tabular}{c|c|c|ccc}
\multicolumn{1}{c|}{\multirow{2}{*}{Teacher}} & \multicolumn{1}{c|}{\multirow{2}{*}{Student}}& \multicolumn{1}{c|}{\multirow{2}{*}{Learnable}}& \multicolumn{3}{c}{$N_{A}$}\\
& & & 2 & 4 & 6\\\hline
\multirow{2}{*}{WRN-40-2} & \multirow{2}{*}{WRN-16-2}& Yes  & 76.75 & 76.58 & 76.55 \\
& &  No & 76.41 & 76.05 & 75.66 \\\hline
\multirow{2}{*}{ResNet110} & \multirow{2}{*}{ResNet20} & Yes  & 72.34 & 72.44 & 72.43\\
 & & No & 71.86 & 71.74 & 71.80\\ 
\end{tabular}}
\caption{Top-1 test accuracy (\%) comparison of whether the neural network-based data augmentation module is learnable or not on CIFAR-100.}\label{tab:iflearningresult}
\end{table}
\begin{table}[t]
	\renewcommand\arraystretch{1.1}
	\setlength\tabcolsep{0.6mm}
	\centering
	\footnotesize
\scriptsize
\setlength{\tabcolsep}{2.8mm}{
\begin{tabular}{c|c|cccc}
\multicolumn{1}{c|}{\multirow{2}{*}{Teacher}}& \multicolumn{1}{c|}{\multirow{2}{*}{Student}}& \multicolumn{4}{c}{$n_\textrm{encoder}$}\\
\multicolumn{1}{c|}{} & & $0\!\cdot\!\frac{|\mathcal{X}|}{B}$ & $1\!\cdot\!\frac{|\mathcal{X}|}{B}$ & $15\!\cdot\!\frac{|\mathcal{X}|}{B}$ & $30\!\cdot\!\frac{|\mathcal{X}|}{B}$\\\hline
WRN-40-2 & WRN-16-2 & 76.05 & 76.72 & 76.53 & 76.58 \\
ResNet110 & ResNet20 & 71.74 & 72.46 & 72.41 & 72.44 \\
\end{tabular}}
\caption{Top-1 test accuracy (\%) comparison on CIFAR-100. Here, we performed ablation experiments on $n_\textrm{encoder}$. $N_{A}$ for all experiments in this table is set as 4.}\label{tab:iterresult}
\end{table}
\begin{table}[t]
	\renewcommand\arraystretch{1.1}
	\setlength\tabcolsep{0.6mm}
	\centering
\footnotesize
\scriptsize
\setlength{\tabcolsep}{2.8mm}{
\begin{tabular}{c|c|cccc}
\multicolumn{1}{c|}{\multirow{2}{*}{Teacher}}& \multicolumn{1}{c|}{\multirow{2}{*}{Student}}& \multicolumn{4}{c}{Mode}\\
\multicolumn{1}{c|}{} & & Attack & Ours$^{+}$ & Ours$^{+*}$ & Ours\\\hline
WRN-40-2 & WRN-16-2 & 74.76 & 66.01 & NAN & 76.58 \\
ResNet110 & ResNet20 & 71.15 & 52.91 & NAN & 72.44 \\
\end{tabular}}
\caption{Top-1 test accuracy (\%) comparison for different data encoders on CIFAR-100. Here, $+$ and $*$ stand for not freezing all the parameters in ${\{f^i_E\}}_{i=1}^N$ and not introducing bias of a priori data augmentation into ${\{f^i_E\}}_{i=1}^N$, respectively. The results ``NAN'' represents the gradient explosion. Besides, $\mathcal{N}_{A}$ and $n_\textrm{encoder}$ for all experiments in this table is set as 4 and $30\!\cdot\!\frac{|\mathcal{X}|}{B}$, respectively.}\label{tab:moderesult}
\end{table}
We conduct ablation studies in three aspects: \textbf{(a)} the effect of Stage II on the student performance; \textbf{(b)} the effect of different iteration numbers in Stage II; \textbf{(c)} the impact of varying data encoders on TST. 

\begin{figure}[ht]
\begin{flushleft}
\includegraphics[width=0.53\textwidth,trim={1.4cm 1.4cm 0cm 0cm},clip]
{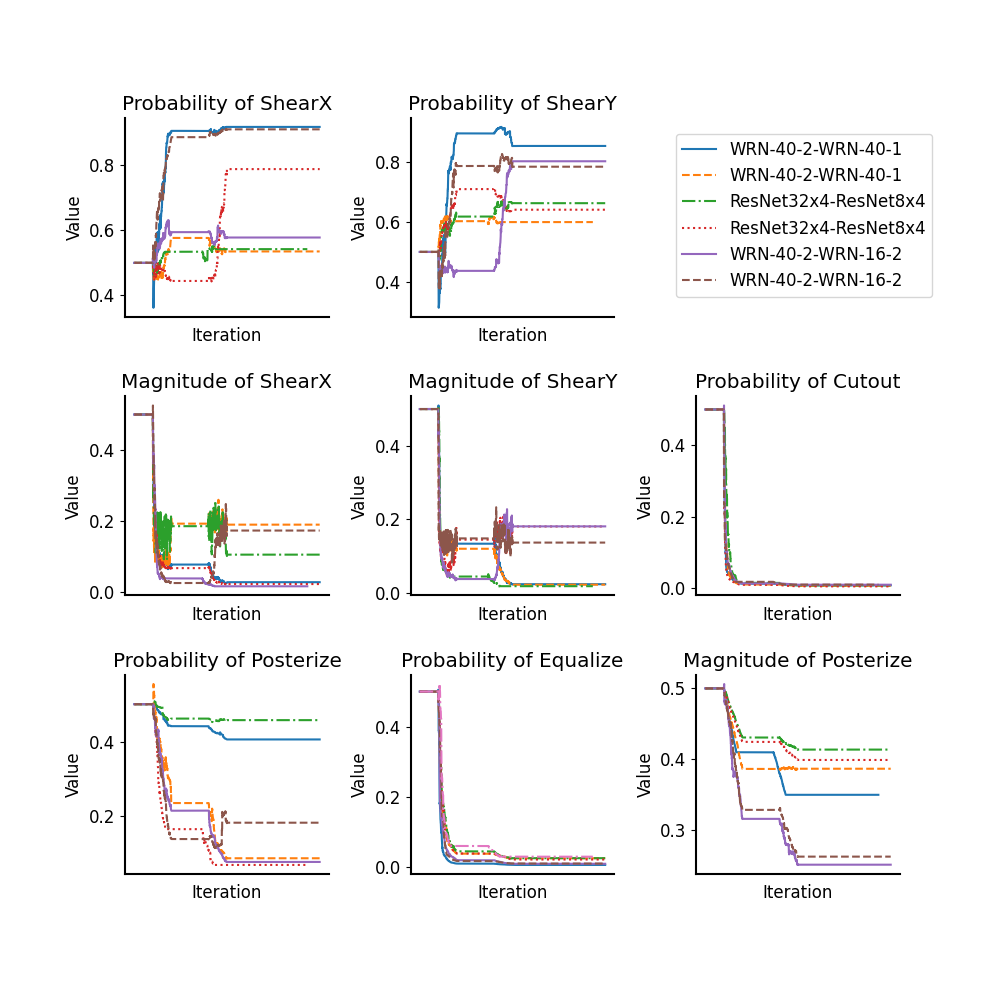}
\end{flushleft}
\caption{The plot of magnitudes and probabilities variation with iteration under TST on CIFAR-100. Here we have selected representative sub-policies, including ShearX, ShearY, Cutout, Posterize, and Equalize.}
\label{fig:mp}
\end{figure}

As illustrated in Table~\ref{tab:iflearningresult}, no matter what kind of teacher-student pairs or whatever value of $N_A$, the augmented sample search strategy of TST, i.e., Stage II, is always beneficial for distillation. 

In addition, we suppose that TST may easily find augmented samples that the student is not good at, but it is difficult for the student to absorb the corresponding knowledge fully. Based on this conjecture, TST should perform best in the case that $n_\textrm{encoder}\ll n_\textrm{student}$. The experimental results in Table~\ref{tab:iterresult} verify our assumption. Note that $\frac{|\mathcal{X}|}{B}$ actually refers to the iteration number within an epoch.

At last, we show the impact of different types of data encoders on the performance of TST in Table~\ref{tab:moderesult} and analyze it. One of the data encoders, Attack, denotes the direct manipulation of samples in a form similar to PGD~\cite{PGD}, which means that TST must let Stage II and Stage III iterate once each in turn and repeat the process. Based on the findings in Table~\ref{tab:moderesult}, we can indicate that it is an extremely sensible choice to introduce the augmented priori bias into the data encoder and to freeze the parameters associated with the bias during distillation process.

\begin{table}[t]
\renewcommand\arraystretch{1.1}
\center
\begin{small}
\scriptsize
\setlength{\tabcolsep}{2.8mm}{
\begin{tabular}{cccc|cc}
\multicolumn{4}{c|}{Hyperparameters} & \multicolumn{2}{c}{Teacher}\\
$w_\textrm{ce}$& $w_\textrm{kl}$ & $\alpha$ & $\beta$  & WRN-40-4 & WRN-28-10 \\\hline
1.0 & 1.0 & 1.00 & 1.00 & 75.93 & 75.79 \\ 
1.0 & 1.0 & 0.95 & 1.05 & 76.13 & 76.04 \\ 
1.0 & 1.0 & 0.85 & 1.15 & 76.35 & 76.18 \\ 
1.0 & 1.0 & 0.75 & 1.25 & 76.42 & 76.34 \\ 
1.0 & 1.0 & 0.5 & 1.5 & 76.45 & 75.80 \\ 
1.0 & 1.0 & 0.25 & 1.75 & 76.19 & 75.81 \\\hline
0.8 & 1.2 & 1.00 & 1.00 & 76.32 & 75.89 \\ 
0.6 & 1.4 & 1.00 & 1.00 & 76.37 & 75.96 \\ 
0.4 & 1.6 & 1.00 & 1.00 & 76.47 & 76.02 \\ 
0.2 & 1.8 & 1.00 & 1.00 & 76.29 & 76.17 \\ 
\end{tabular}}
\end{small}
\caption{Top-1 accuracy (\%) comparison on CIFAR-100. Analytical experiments of $w_\textrm{ce}$, $w_\textrm{kl}$, $\alpha$ and $\beta$ hyperparameters. Here, all students are WRN-16-2.}
\label{tab:weaknessresult}
\end{table}
\section{Discussion}
\paragraph{How to alleviate the gap between the teacher and student?} To analyze this situation, we train two stronger teachers, including WRN-40-4 and WRN-28-10, achieve 80.7\% and 82.0\% Top-1 accuracy on the test set of CIFAR-100, respectively. Then, we employ these two teachers for a more in-depth exploration. As illustrated in Table~\ref{tab:weaknessresult}, when $w_\textrm{ce}$, $w_\textrm{kl}$, $\alpha$ and $\beta$ take default values (i.e., 1, 1, 1 and 1), stronger teachers (i.e., WRN-40-4 and WRN-28-10) usually achieve worse performance compared to WRN-40-2. To analyze this phenomenon, we show in Fig.~\ref{fig:confidence} that the teacher's response to the ground truth label, i.e., the probabilities of correctly discriminating samples. Specifically, in each epoch of Stage III, we get the teacher's responses to the correct category corresponding to all training samples, and plot the expectation of the responses. This indicates that the stronger the teacher is, the closer its response is to one-hot encoding and the less ``dark knowledge'' attached to the logit. Consequently, the performance degradation of students caused by stronger teachers can be mitigated by making the teacher logit not too close to one-hot coding. We propose two solutions: (a) decrease $w_\textrm{ce}$ and increase $w_\textrm{kl}$, simultaneously, to transfer more ``dark knowledge'' from the teacher to the student; (b) decrease $\alpha$ and increase $\beta$, simultaneously, to make the teacher not overconfident about the augmented samples searched by Stage II. As exhibited in Table~\ref{tab:weaknessresult}, both methods assist in bridging the gap between the teacher and student to some extent.

\paragraph{What are the strengths and probabilities of execution for augmentations that the teacher excels at, but the student does not?} To answer this question, we choose 8 magnitudes and probabilities from $\{\theta_m^i\}_{i=1}^{N-N_\textrm{nl}}$ and $\{\theta_p^i\}_{i=1}^{N}$ with distinct patterns and plot their variation with iteration in Fig.~\ref{fig:mp}. The full TST visualization of search results is shown in Appendix~\ref{apd:sd}. Note that there are 3 teacher-student pairs, i.e., WRN-40-2-WRN-16-2, WRN40-2-WRN-40-1 and ResNet32$\times$4-ResNet8$\times$4 pairs in these figures, and each pair has two changing curves. Although the magnitudes and probabilities searched by the different models are various, they have macroscopic similarities. For instance, ShearX and ShearY converge to larger probabilities and smaller magnitudes; sub-policies without learnable magnitudes, such as Equalize and Cutout, converge to minimal probabilities. In fact, such a result is intuitive. For instance, the co-characteristics of the dataset let the magnitudes and probabilities searched by the data augmentation methods (e.g., PBA and AutoAugment) on one model can also be applied to another one. Besides, each curve in Fig.~\ref{fig:mp} differs at a microscopic level due to the difference between teacher-student pairs and the stochastic nature of deep learning.
\section{Conclusion}
In this paper, inspired by realistic teaching scenarios, we design a data-based distillation algorithm called TST. To be specific, TST locates augmented samples that the teacher is good at while the student is not, and transfers the knowledge of these augmented samples from the teacher to the student with the expectation that an excellent student can be learned. Our experimental results, including both qualitative and quantitative ones, demonstrate the feasibility of the data-based distillation method and the validity of the ``Teach what you Should Teach'' strategy. In future research, the data-based distillation, which is as competitive as logit-based and feature-based distillation, may deserve more attention in the field of knowledge distillation. Meanwhile, combining data-based distillation with logit-based distillation and feature-based distillation may lead to greater performance breakthroughs in distillation.

\section*{Acknowledgments} This work was supported in part by the Nature Science Foundation of China (NSFC) under Grant No 62072041.

{\small
\bibliographystyle{named2023}
\bibliography{egbib}
}
\clearpage
\clearpage
\appendix
\section{Datasets}
\label{apd:dataset}
\paragraph{CIFAR-100.} Dataset CIFAR-100~\cite{CIFAR} is the subsets of the tiny image dataset and consists of 60,000 images with the size 32$\times$32. Specifically, the training set contains 50,000 images, and the testing set contains 10,000 images. 
\paragraph{ImageNet-1k.} Dataset ImageNet-1k~\cite{ILSVRC15}, also commonly referred to as ILSVRC 2012, has 1000 classes, and the benchmark is trained using the training set and tested using the validation set. Its training and validation sets contain 1281,167 and 50,000 images, respectively.
\paragraph{MS-COCO.} Dataset MS-COCO~\cite{COCO} is a large-scale object detection dataset. The benchmark is the same as ImageNet-1k, using the training set for training and the validation set for testing. The training/validation split was changed from 83K/41K to 118K/5K in 2017. Researchers commonly apply the 2017 version for experiments.
\paragraph{Cityscapes.} Dataset Cityscapes~\cite{cordts2016cityscapes} is a new large-scale dataset for semantic segmentation. It provides 5,000 images that have been meticulously annotated, with 2,975 images for training and 500 images for validation, where 30 common classes are provided and 19 classes are used for evaluation and testing. Each image is 2048$\times$1024 in size.

\section{Hyperparameter Settings}
\label{apd:hsettings}
\subsection{Basic settings} 
\paragraph{Classification.} For the classification experiments on CIFAR-100, the batch size is 128, the total number of epochs is 240, and the learning rate is initialized to 0.1 and divided by 10 at 150, 180 and 210 epochs. In addition, we employ an SGD optimizer for training and set the weight decay and momentum as 5e-4 and 0.9, respectively. For the classification experiments on ImageNet-1k (ResNet34-ResNet18 pair and ResNet50-MobileNet pair), the total batch size is 512, the total number of epochs is 100, the batch size in every GPU is 128, the number of GPUs is 4 and the learning rate is initialized to 0.1 and divided by 10 at 30, 60 and 90 epochs. Besides, we employ an SGD optimizer for training and set the weight decay and momentum as 1e-4 and 0.9, respectively. For the Swin-Large-Swin-Tiny pair, we follow the setting in~\cite{SWIN-T}, except for the batch size and initial learning rate. Specifically, the batch size is 512, the total number of epochs is 300, the batch size in every GPU is 64, the number of GPUs is 8, and the learning rate is initialized to 0.0005 and is decayed by a cosine scheduler. Furthermore, we utilized an Adam optimizer for training and set the weight decay as 5e-2. The reason for halving the batch size is that the GPU memory is insufficient to support the training with the original hyperparameter settings.

\paragraph{Detection.} For the detection experiments on MS-COCO, we utilize mmdetection~\cite{mmdetection} and mmrazor~\cite{2021mmrazor} for both training and testing. Following~\cite{CWD,RKD}, we use the same standard training strategies on the Cascade RCNN-X101-Faster RCNN-R50 and RetinaNet-X101-RetinaNet-R50 pairs. To be specific, the total batch size is 16, the total number of epochs is 24, the batch size in every GPU is 2, the number of GPUs is 8 and the learning rate is divided by 10 at 16 and 22 epochs. The initial learning rate is set as 0.02 and 0.01 on Cascade RCNN-X101-Faster RCNN-R50 and RetinaNet-X101-RetinaNet-R50 pairs, respectively. Besides, the setting on the FCOS-R101-FCOS-R50 pair is following~\cite{FGD}. Compared with the RetinaNet-X101-RetinaNet-R50 pair, the only difference is we apply a warm-up learning rate on the FCOS-R101-FCOS-R50 pair.

\paragraph{Segmentation.} For the segmentation experiments on Cityscapes, we apply mmsegmentation~\cite{mmseg2020} and mmrazor~\cite{2021mmrazor} for distillation. we follow the setting in~\cite{2021mmrazor}. In specific, the total batch size is 16, the total number of iterations is 80,000, the batch size in every GPU is 2, the number of GPUs is 8 and the learning rate is 0.01. We make the learning rate decay to 0.9 in each iteration and constrain the minimum learning rate in training to be 1e-4. And we utilized a SGD optimizer for training and set the weight decay and momentum as 5e-4 and 0.9, respectively.

\subsection{Advanced Settings}
\paragraph{Classification.} On CIFAR-100, the loss weight is set as 1 for all comparison results and the temperature $\tau$ of TST (i.e., vanilla KD) is 4. We let $f_{DE}$ update the magnitudes $\{\theta_m^i\}_{i=1}^{N-N_\textrm{nl}}$ and probabilities $\{\theta_p^i\}_{i=1}^{N}$ (i.e., conduct Stage II) at 30 and 90 epochs and continuously train $f_{DE}$ with ALRS~\cite{chen2022bootstrap} for 1 epoch. Moreover, on ImageNet-1k, the loss weight is set as 1 for all comparison results and the temperature $\tau$ of TST (i.e., vanilla KD) is 1. The magnitudes $\{\theta_m^i\}_{i=1}^{N-N_\textrm{nl}}$ and probabilities $\{\theta_p^i\}_{i=1}^{N}$ are updated at 10 and 20 epochs. Unlike CIFAR-100 training, we train 5 epochs for $f_{DE}$ continuously.

\paragraph{Detection.} On MS-COCO, the loss weight is set as 1 for all comparison results and the temperature $\tau$ of TST (i.e., vanilla KD) is 1. We add $\mathcal{L}_{KL}$ and $\mathcal{L}_{MSE}$ to the final predictions of classes and the neck, respectively. On Stage II, we let not only the student classification predictions but also the student box regression be inaccurate. Compare with the student, we make the teacher's categorical predictions and box regressions as accurate as possible. Furthermore, We make $f_{DE}$ update the magnitudes $\{\theta_m^i\}_{i=1}^{N-N_\textrm{nl}}$ and probabilities $\{\theta_p^i\}_{i=1}^{N}$ at 6 and 12 epochs and continuously train $f_{DE}$ for 1 epoch. 

\paragraph{Segmentation.} On Cityscapes, we set the loss weight on $\mathcal{L}_{KL}$ between the teacher decode head's final predictions and the student decode head's final predictions as 1.5. In addition, we set the loss weight on $\mathcal{L}_{KL}$ between the teacher auxiliary head's final predictions and the student auxiliary head's final predictions as 0.5. Also, the temperature $\tau$ of TST (i.e., vanilla KD) is set as 1. Last but not least, the magnitudes $\{\theta_m^i\}_{i=1}^{N-N_\textrm{nl}}$ and probabilities $\{\theta_p^i\}_{i=1}^{N}$ is updated at 7,500 and 15,000 iterations on the total training. We continuously update $\{\theta_m^i\}_{i=1}^{N-N_\textrm{nl}}$ and probabilities $\{\theta_p^i\}_{i=1}^{N}$ with 1,250 iterations.

\section{Training Process on TST}
\label{apd:TST_procedure}
In this section, we show the full training process of TST in Algorithm~\ref{alg:ta}. For the sake of simplicity, we do not categorize the different sub-policies here and consistently consider that they are all learnable. Moreover, for the object detection task, Eq.~\ref{eq:totalloss} in Algorithm\ref{alg:ta} can be replaced with other forms, i.e., $\mathcal{L}_\textrm{mimic}+\mathcal{L}_\textrm{KL}+\mathcal{L}_\textrm{CE}$.
\begin{algorithm}[t]
\caption{Training procedure of TST}
\label{alg:ta}
\begin{algorithmic}[1]
\renewcommand{\algorithmicrequire}{\textbf{Input:}}
\renewcommand{\algorithmicensure}{\textbf{Output:}}
\REQUIRE A student $f_S$ with the parameter $\theta_S$, a teacher model $f_T$ with the parameter $\theta_T$, a data augmentation encoder set ${\{f^i_E\}}_{i=1}^N$ with the parameter set ${\{\theta^i_E\}}_{i=1}^N$, a set of the priori manual data-augmentation mappings ${\{f^i_A\}}_{i=1}^N$, the parameter of probabilities $\theta_p$, the parameter of magnitudes $\theta_m$, dataset $\mathcal{X}$, the probabilistic transformation function of the ground truth label $p$, the number of iterations required for the priori manual data augmentation fitting $n_\textrm{fitting}$, the number of iterations required for a encoder training phase $n_\textrm{encoder}$, the number of iterations required for a student training phase $n_\textrm{student}$, the iterative position set $\mathbb{I}$ for probability and magnitude learning, the learning rate $\eta_{\theta_S}$ and $\eta_{\theta_E}$.
\STATE Random initialization parameter $\{\theta^i_E\}_{i=1}^N$.
\FOR{$i=1,\cdots,N$}
\FOR{$j=0,\cdots,n_\textrm{fitting}$}
\STATE Randomly sample a mini-batch, $\{x^k\}^B_{k=1}\sim\mathcal{X}$.
\STATE Randomly sample the magnitude $m$.
\STATE $\{\ve{y}^{i,k}_{E}\}_{k=1}^B = \{f^i_E\left(x^k,m\right)\}_{k=1}^B$.
\STATE $\{\ve{y}^{i,k}_{A}\}_{k=1}^B = \{f^i_A\left(x^k,m\right)\}_{k=1}^B$.
\STATE Compute loss for the data augmentation encoder,\\$\mathcal{L}^i_\textrm{encoder}=\sum_{k=1}^B\|\ve{y}^{i,k}_{E}-\ve{y}^{i,k}_{A}\|_2$.
\STATE Update $\theta^i_E$ by the gradient ascent,\\$\theta^i_E\leftarrow \theta^i_E -\eta_{\theta_E}\partial \mathcal{L}^{i}_\textrm{encoder}/\partial\theta^i_E$.
\ENDFOR
\ENDFOR
\STATE Random initialization parameter the parameter of probabilities $\theta_p$ and the parameter of magnitudes $\theta_m$.
\FOR{$i=0,\cdots,n_\textrm{student}$}
\IF{$i \in \mathbb{I}$}
\FOR{$j=0,\cdots,n_\textrm{encoder}$}
\STATE Randomly sample a mini-batch, $\{x^k\}^B_{k=1}\sim\mathcal{X}$.\STATE Calculate new samples $\{\hat{x}^k\}^{B}_{k=1}$ using Eq.~\ref{eq:total_nbdam_forward}.
\STATE Compute loss for probabilities and magnitudes,\\ $\mathcal{L}_\textrm{TST}=\sum_{k=1}^{B}(-\log(p(f_S(\hat{x}^k)))-\log(1-p(f_T(\hat{x}^k))))$
\STATE Update $\phi$ by the gradient ascent, $\theta_p,\theta_m\leftarrow \theta_p-\eta_\phi\partial \mathcal{L}_\textrm{TST}/\partial\theta_p,\theta_m-\eta_\phi\partial \mathcal{L}_\textrm{TST}/\partial\theta_m$
\ENDFOR
\ENDIF
\STATE Randomly sample a mini-batch, $\{x^k\}^B_{k=1}\sim\mathcal{X}$
\STATE Calculate new samples $\{\hat{x}^k\}^{2B}_{k=1}$ by Eq.~\ref{eq:total_nbdam_forward} and the concatenation operator.
\STATE Compute loss $\mathcal{L}_{S}$ for the student by Eq.~\ref{eq:totalloss}.
\STATE Update $\theta_S$ by the gradient descent,\\
$\theta_S\leftarrow \theta_S-\eta_{\theta_S}\partial \mathcal{L}_{S}/\partial\theta_S$
\ENDFOR
\end{algorithmic}
\end{algorithm}

\section{Supplementary Details}
\label{apd:sd}
\subsection{Details about $f_{DE}$ in object detection tasks} 
Generally, a target object, after affine transformation and accumulation according to Eq.~\ref{eq:total_nbdam_forward}, generates multiple target objects, leading to ambiguity in the target detection task. Thus, we design a method to replace the weighted summation of the difference between the affine transform result and the original image. To be specific, we sum all affine transformation matrics generated from $f_\textrm{STN}$ for obtaining only one new target affine transformation matrix $\hat{A}$. And $f_{DE}$ simply learns $\hat{A}$ that satisfies the objective that TST search. In here, we define two hyperparameters $N_\textrm{a}$ and $N_\textrm{c}$ (s.t., $N_\textrm{a}+N_\textrm{c} = N_A $) refer to the number of learnable affine transformations and color transformations that are actually selected, respectively. Due to all unlearnable transformations are color transformations, we can disregard them. First, for a set of affine meta-encoders $\{f_\textrm{STN}^{Z^i}\}_{i=1}^{N_\textrm{a}}$ (s.t., $N_\textrm{a}\!\leq\!N_\textrm{A}$)\footnote{Here, for simplicity, we let the first $N_\textrm{a}$ of $N_\textrm{A}$ actual applied meta-encoders all be $f_\textrm{STN}$.}, we can define the set of affine transformation matrices generated by this set as $\{A^{Z^i}\}_{i=1}^{N_\textrm{a}}$. Then, the finial matrix $\hat{A}$ generated by $f_{DE}$ can be formulated as
\begin{equation}
\begin{aligned}
&A^i_E\!=\theta_p^{Z^i}\odot A^{Z^i}-\theta_p^{Z^i}\odot I,\ \mathop{s.t.}\ i\leq N_\textrm{a}\\
&\hat{A}\!=\!I+\sum_{i=1}^{N_\textrm{a}}A^i_E, \\
\end{aligned}
\label{eq:total_nbdam_forward_2}
\end{equation}
where $I$ refers to a matrix$\sim\mathbb{R}^{2\times3}$ where only the values on the diagonal are 1 and the rest are 0. After that. we derive $\hat{x}_a$ by the operator $\mathbf{\mathrm{Affine}}$. 
\begin{equation}
\begin{aligned}
&\hat{x}_a = \mathbf{\mathrm{Affine}}(\hat{A},x)\\
\end{aligned}
\label{eq:total_nbdam_forward_3}
\end{equation}
Now, we need to apply the remaining meta-encoders to complete the transform of $\hat{x}_a$. As a result, the corresponding computational form is shown in Eq.~\ref{eq:total_nbdam_forward_4}.
\begin{equation}
\begin{aligned}
&x^i_E\!=\begin{cases}\theta_p^{Z^i}\odot f_E(\hat{x}_a,\theta_m^{Z^i})-\theta_p^{Z^i}\odot \hat{x}_a,& \\
N_\textrm{a}<i\leq N_A\ \textrm{and}\ Z^i \leq N_\textrm{nl} &\\
\theta_p^{Z^i}\odot f_E(\hat{x}_a)-\theta_p^{Z^i}\odot \hat{x}_a, & \\
 N_\textrm{a}<i\leq N_A\ \textrm{and}\ Z^i >N_\textrm{nl} &\\
\end{cases}\\
&\hat{x}\!=\!\hat{x}_a+\sum_{i=N_\textrm{a}+1}^{N_A}x^i_E, \\
\end{aligned}
\label{eq:total_nbdam_forward_4}
\end{equation}
Finally, we obtain $\hat{x}$, which is a part of the input to $f_T$ and $f_S$. In particular, $\hat{x}$ in Eq.~\ref{eq:total_nbdam_forward} and Eq.~\ref{eq:total_nbdam_forward_4} have the same meaning.

\subsection{Full probability and magnitude visualization}
\begin{figure*}[t]
\centering
\includegraphics[width=1\textwidth,trim={4cm 2cm 4cm 2cm},clip]{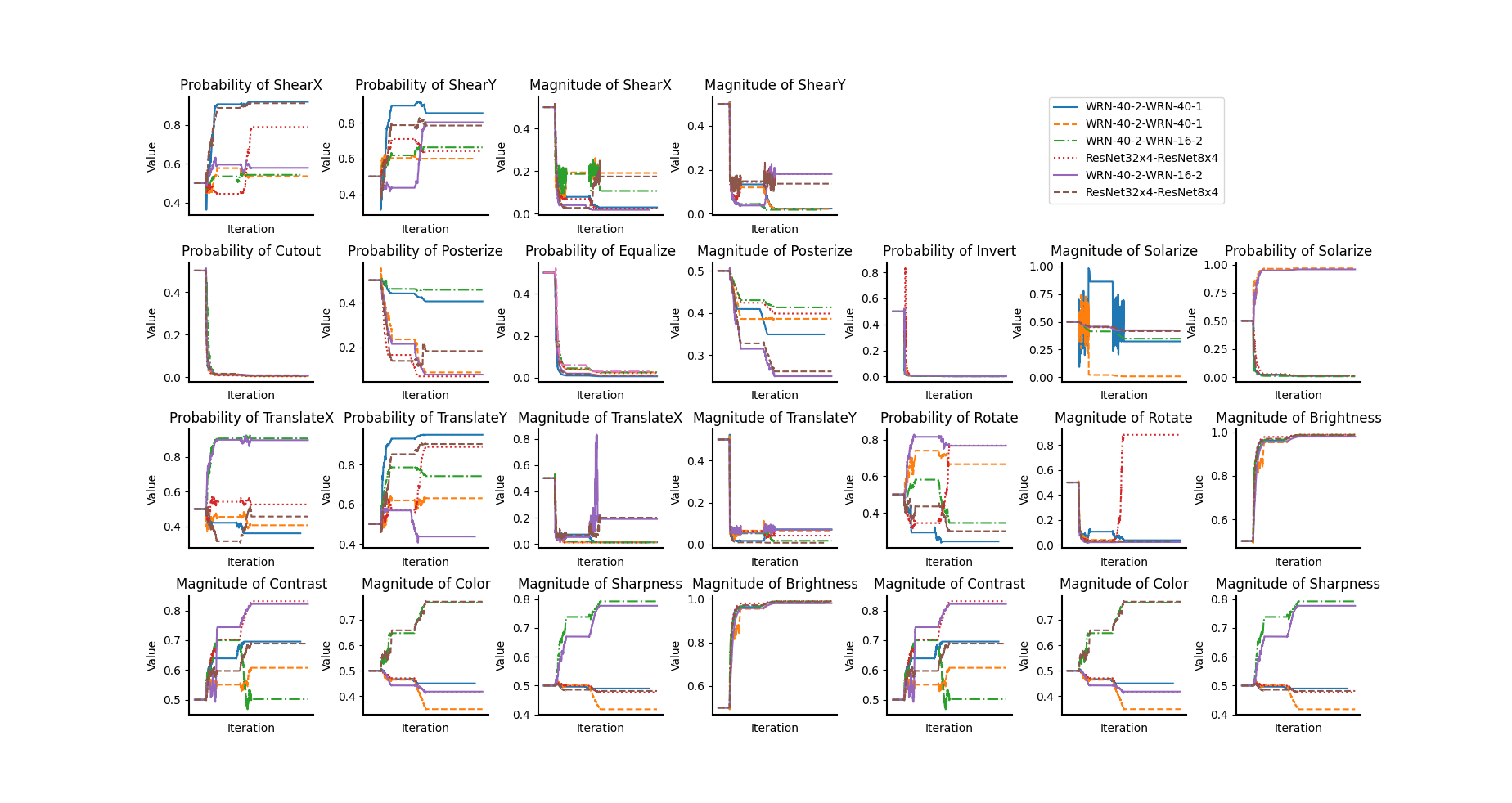}
\caption{The plot of magnitudes and probabilities variation with iteration under TST on CIFAR-100. Here we visualize all sub-policies mentioned in this paper.}
\label{fig:total_mp}
\end{figure*}

Due to space constraints in the main paper, we only show 8 sub-policies' magnitudes and probabilities searched by TST. So, we illustrate the magnitudes and probabilities of all sub-policies in Fig.~\ref{fig:total_mp} (i.e., 11 magnitudes and 14 probabilities).

\subsection{Proof} 
Here, we will prove $-\mathcal{L}_\textrm{CE}\left(x,y\right)$ is a non-convex function and $\mathcal{L}_\textrm{CE}\left(1\!-\!x,y\right)$ is a convex function, where $x$ is the logits output from the normalized activation function, i.e., softmax. And $y$ is the ground truth label. Besides, the lengths of vectors $x$ and $y$ are the number of categories. Since $y$ is one-hot encoding, $-\mathcal{L}_\textrm{CE}\left(x,y\right)$ can be written as $\mathbf{\mathrm{log}}(x_t)$, where $x_t$ is the predicted logit in $x$ corresponding to the target label. In particular, $0\leq x_t \leq 1$ due to the presence of softmax. As the definition of a convex function, we need to prove the following equation (note that deep learning is stochastic gradient descent for parameter updating, so the convex function declared here is downward convex):
\begin{equation}
\small
\begin{aligned}
\mathbf{\textrm{log}}(a+b) \leq \mathbf{\textrm{log}}(a) + \mathbf{\textrm{log}}(b), \quad \mathop{s.t.}\quad 0\leq a\leq b \leq 1.
\end{aligned}
\label{eq:non_convex_log}
\end{equation}
Obviously, $a+b\geq 2\sqrt{ab} \geq ab$, so Eq.~\ref{eq:non_convex_log} does not hold and $-\mathcal{L}_\textrm{CE}\left(x,y\right)$ is a non-convex function. Similarly, for $\mathcal{L}_\textrm{CE}\left(1\!-\!x,y\right)$, we can infer that it is a convex function by Eq.~\ref{eq:convex_log} since $ab\leq 1$ is a known condition.
\begin{equation}
\small
\begin{aligned}
& -\mathbf{\textrm{log}}(2-a-b) \leq -\mathbf{\textrm{log}}(1-a) - \mathbf{\textrm{log}}(1-b), \\
& \quad \mathop{s.t.}\quad 0\leq a\leq b \leq 1, \\
&=>\frac{1}{2-a-b} \leq \frac{1}{(1-a)(1-b)},\\
&=>ab\leq 1,\\
\end{aligned}
\label{eq:convex_log}
\end{equation}
\begin{table}[h]
\vspace{-8pt}
\renewcommand\arraystretch{1.1}
\center
\caption{Top-1 test accuracy (\%) comparison of the few-shot scenario on CIFAR-100. Here, all teacher-student pairs are ResNet56-ResNet20 pairs.}
\label{tab:fewshotresult}
\scriptsize
\setlength{\tabcolsep}{2.8mm}{
\begin{tabular}{c|ccc}
\multicolumn{1}{c|}{Percentage} & KD & CRD & Ours \\\hline
25\% & 65.15 & 65.80 & \textbf{66.28} \\
50\% & 68.61 & 69.91 & \textbf{70.22} \\
75\% & 70.34 & 70.68 & \textbf{71.80} \\
\end{tabular}}
\vspace{-8pt}
\end{table}
\begin{figure}[t]
\centering
\includegraphics[width=0.3\textwidth,height=0.2\textheight]{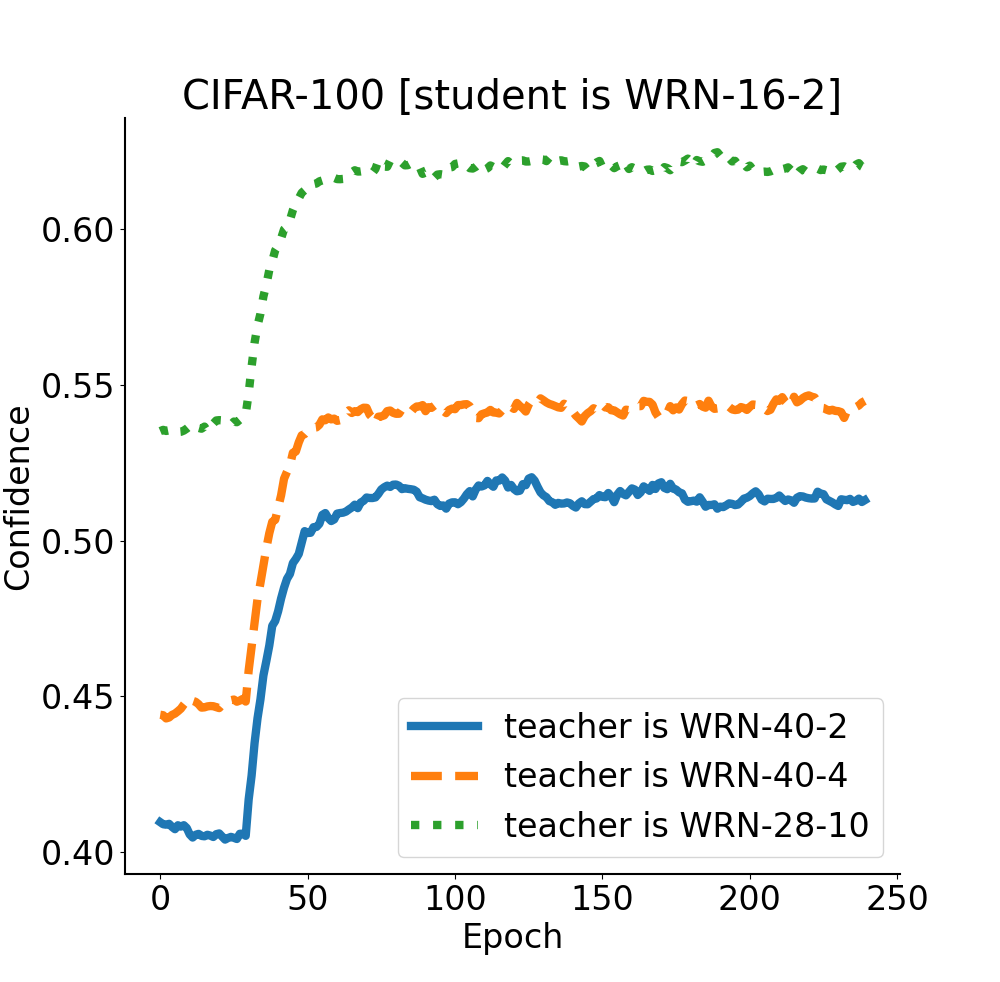}
\caption{The probability of different-size teachers predicting samples correctly with epochs. Here, this figure is intended to supplement the discussion section.}
\label{fig:confidence}
\end{figure}

\subsection{Combine DIST and TST}
TST is a novel data-based distillation algorithm that can be combined with other arbitrary feature-based and logit-based distillation algorithms and further enhance students' generalization ability. For instance, we combine DIST and TST on ResNet50-MobileNet-V1 on ImageNet-1k with 33 hours, which boosts the results from 73.24\% to 73.55\%.

\subsection{Generative Adversarial Network and Learnable Data Augmentation}
\label{sec:gan_and_mda}
We started by encoding the original samples based on Generative Adversarial Networks, but without some regularization constraints directly causing training collapse, and even with some constraints, the performance (75.60\%) is still not as good as the current system, i.e., learnable data augmentation (76.75\%) on WRN-40-2-WRN-16-2 pair on CIFAR-100.

\subsection{Supplementary Materials}
\label{apd:few_shot_experiment}
Due to space constraints in the text, we present here the comparison experiments in the few-shot scenario (Table~\ref{tab:fewshotresult}) and an iterative plot of the teacher's output of the logit's confidence as a function of epoch (Table~\ref{fig:confidence}).

\end{document}